\pdfoutput=1
\documentclass[12pt]{article}

\usepackage{sbc-template}

\usepackage{graphicx}
\usepackage[utf8]{inputenc}
\usepackage{amsmath}
\usepackage{booktabs}
\usepackage{multirow}
\usepackage[colorlinks=true,
            linkcolor=black,
            citecolor=black,
            urlcolor=black,
            filecolor=black,
            breaklinks=true]{hyperref}
\hypersetup{pdftitle={Integrating Simulation and Machine Learning for Autonomous Aerial Combat Agents},
            pdfauthor={Jo\~ao Paulo de Andrade Dantas, Marcos R. O. A. Maximo, Takashi Yoneyama}}

\usepackage{xurl}
\usepackage[most]{tcolorbox}

\title{Integrating Simulation and Machine Learning for Autonomous Aerial Combat Agents}

\author{João Paulo de Andrade Dantas\inst{1}, Marcos R. O. A. Maximo\inst{1}, Takashi Yoneyama\inst{1}}

\address{Instituto Tecnológico de Aeronáutica (ITA)\\
  São José dos Campos, SP, Brazil
  \email{\{jpdantas,mmaximo,takashi\}@ita.br}}

\begin{document}

\maketitle

\begin{abstract}
This work presents a research framework that brings together detailed simulation and machine learning to improve decision support and autonomy for aerial vehicles, with beyond-visual-range (BVR) air combat as a specific case study in the air domain. The framework covers five connected areas: a simulation environment and services; learning-based models of engagement zones and kill probability; in-flight tactical decision support for launch timing and situational awareness; tools for aerospace data analysis; and autonomous agents trained through reinforcement, imitation, and generative learning. This study focuses on the last area: agents learn cooperative tactics from reward and reproduce expert maneuvers from a small set of demonstrations. Although developed for air combat, the methodology can be applied to other applications in aviation, such as commercial flight operations and pilot training.
\end{abstract}


\noindent\textbf{Level:} Ph.D. \quad \textbf{Thesis conclusion:} 27 June 2025.\\
\noindent\textbf{Advisor:} Takashi Yoneyama. \quad \textbf{Co-advisor:} Marcos R. O. A. Maximo.
\vspace{6pt}

\section{Introduction}
\label{sec:intro}

Autonomy is a central theme in modern robotics, and aerial vehicles are among its most demanding platforms: they must perceive, decide, and act in environments that are fast, dynamic, partially observable, and often adversarial. In this setting, machine learning supports stronger situational awareness, faster decision-making, and agents able to operate alone or alongside human pilots~\cite{dantas2024loyal}.

Like most learning-based robots, these agents need large amounts of interaction data, which is scarce and costly to gather on real aircraft, and, in the defense context, often classified. Simulation is the standard answer: a safe, flexible, and low-cost environment to generate scenarios and to train, test, and validate models before deployment~\cite{costa2025simulation}. What makes it usable is fidelity, since the closer the simulated aircraft, sensors, and weapons are to the real ones, the more of what an agent learns survives the transfer.

Despite steady progress, simulation tools, weapon models, tactical decision aids, data-analysis pipelines, and autonomous agents are usually studied in isolation~\cite{costa2025simulation}. The same fragmentation appears inside the agent literature: reinforcement learning (RL) for beyond-visual-range (BVR) air combat is dominated by one-versus-one engagements~\cite{hu2021application,zhang2022maneuver}, and work on larger formations still rewards each agent individually, so the reward carries no information about the ally~\cite{liu2022multi,zhou2024cooperative}. On the imitation side, behavior cloning has reproduced piloting skills~\cite{sandstrom2022fighter} and generative models such as variational autoencoders can enlarge small datasets~\cite{kingma2013auto}, without a systematic account of when that augmentation helps. This motivates the central question of the doctoral thesis summarized in this paper~\cite{dantas2025simulation}:

\begin{tcolorbox}[colback=gray!20, colframe=gray!60, title={\centering \textbf{Research Question}}]
\small How can a modern air force integrate simulation and machine learning to support decision-making and autonomy in air combat operations, and which areas are most critical to enabling this integration?
\end{tcolorbox}

The thesis answers it with a unified framework built on five connected areas, described in Section~\ref{sec:framework}. BVR air combat is the case study throughout: the dominant form of modern aerial engagement~\cite{costa2025simulation}, and a challenging robotics problem, being multi-agent, partially observable, and adversarial under tight time constraints. The main contribution is the framework and the solutions developed within each area, all peer-reviewed and published, as summarized below.

\begin{tcolorbox}[colback=gray!20, colframe=gray!60, title={\centering \textbf{Doctoral Research Outputs}}]
\small The doctoral research produced \textbf{12 journal articles}, all accepted for publication, \textbf{6 of them in journals ranked above the 90th percentile} in their fields (Scopus), and \textbf{17 peer-reviewed conference papers}\footnotemark[1]. The journals include Scientific Data (Nature Portfolio)~\cite{patrikar2025image}, IEEE Robotics and Automation Letters~\cite{navarro2024sorts}, IEEE Access~\cite{costa2025simulation,dantas2025deep,dantas2025autonomous,silva2025asafg}, the Journal of Defense Modeling and Simulation~\cite{dantas2025poker}, and IEEE Latin America Transactions~\cite{dantas2022machine}; the conferences include the ACM SIGSIM Conference on Principles of Advanced Discrete Simulation (SIGSIM-PADS)~\cite{dantas2023autonomous,dantas2024asapy,dantas2024loyal}, the IEEE Winter Simulation Conference (WSC)~\cite{dantas2022supervised,dantas2024autonomous}, the Latin American Robotics Symposium and Brazilian Symposium on Robotics (LARS/SBR)~\cite{dantas2021engagement}, and the Aerial Robotics Workshop of the IEEE International Conference on Robotics and Automation (ICRA)~\cite{patrikar2022challenges}, among others. The research also delivered \textbf{7 software artifacts}, including \textbf{3} registered with the Brazilian National Institute of Industrial Property (INPI), \textbf{1} registered at Carnegie Mellon University, and \textbf{2} released as open source. Part of the research was carried out during a 14-month research exchange at the Robotics Institute, Carnegie Mellon University (November 2021 to January 2023).
\end{tcolorbox}
\footnotetext[1]{Further information about the first author is available at \url{https://www.joaopadantas.com}.}

This paper examines the fifth area, aerial autonomous agents, through learning from reward (Section~\ref{sec:drl}) and from demonstrations (Section~\ref{sec:imitation}); Section~\ref{sec:discussion} discusses lessons and limitations, and Section~\ref{sec:conclusion} concludes.

\section{Research Framework}
\label{sec:framework}

\begin{figure}[ht]
\centering
\includegraphics[width=0.63\textwidth]{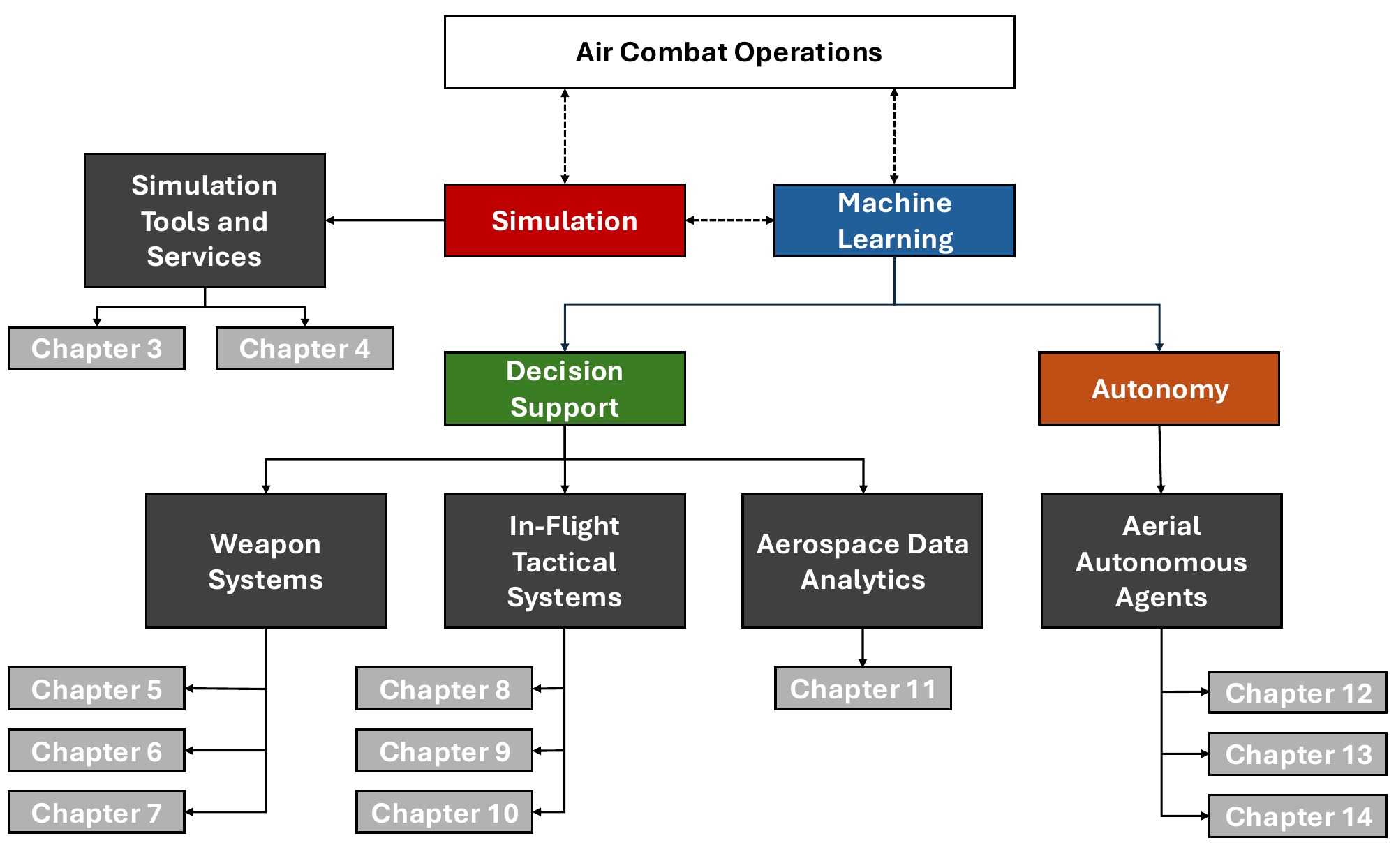}
\caption{The five areas of development. The numbered boxes refer to the chapters of the thesis~\cite{dantas2025simulation}.}
\label{fig:framework}
\end{figure}

The framework organizes the problem into five areas, shown in Figure~\ref{fig:framework}. Simulation, machine learning, and air combat operations sit at the same level and form a cycle in which progress in each one feeds the others. Below them, the areas split into two branches, decision support and autonomy.

\textbf{(I) Simulation tools and services.} The Aerospace Simulation Environment (ASA), developed by the Institute for Advanced Studies (IEAv) of the Brazilian Air Force, simulates aircraft, radars, radar warning receivers, datalinks, and missiles~\cite{dantas2025asa}, and is exposed as a set of services so large batches of scenarios run on demand~\cite{dantas2025asasimaas}, while a human-in-the-loop module lets pilots fly in the same scenarios~\cite{silva2025asafg}. Each aircraft is driven by a Behavior Tree (BT) encoding standard combat behaviors~\cite{colledanchise2020behavior}.

\textbf{(II) Weapon systems.} Learned models estimate the Weapon Engagement Zone (WEZ) of air-to-air missiles~\cite{dantas2021weapon}, predict surface-to-air engagement zones in real time~\cite{dantas2023real}, and estimate probability of kill against stochastic targets~\cite{dantas2025poker}.

\textbf{(III) In-flight tactical systems.} Supervised models support decisions during the engagement, including missile launch timing and situational awareness from partial sensor information~\cite{dantas2021engagement,dantas2022supervised,dantas2022machine}.

\textbf{(IV) Aerospace data analytics.} AsaPy is a Python library for design of experiments, execution control, and analysis of large simulation datasets~\cite{dantas2024asapy}.

\textbf{(V) Aerial autonomous agents.} Agents learn air combat behavior in two ways: RL inside AsaGym, a library that wraps ASA as a standard learning environment~\cite{dantas2023autonomous,dantas2025deep}, and imitation learning from recorded pilot flights, with generative models synthesizing further trajectories~\cite{dantas2024autonomous,dantas2025autonomous}.

\section{Learning Tactical Behavior with Deep Reinforcement Learning}
\label{sec:drl}

The first approach asks an agent to discover tactical behavior on its own, from interaction and reward~\cite{dantas2023autonomous,dantas2025deep}.

\subsection{Environment and Setup}

\begin{figure}[tb]
\centering
\includegraphics[width=0.72\textwidth]{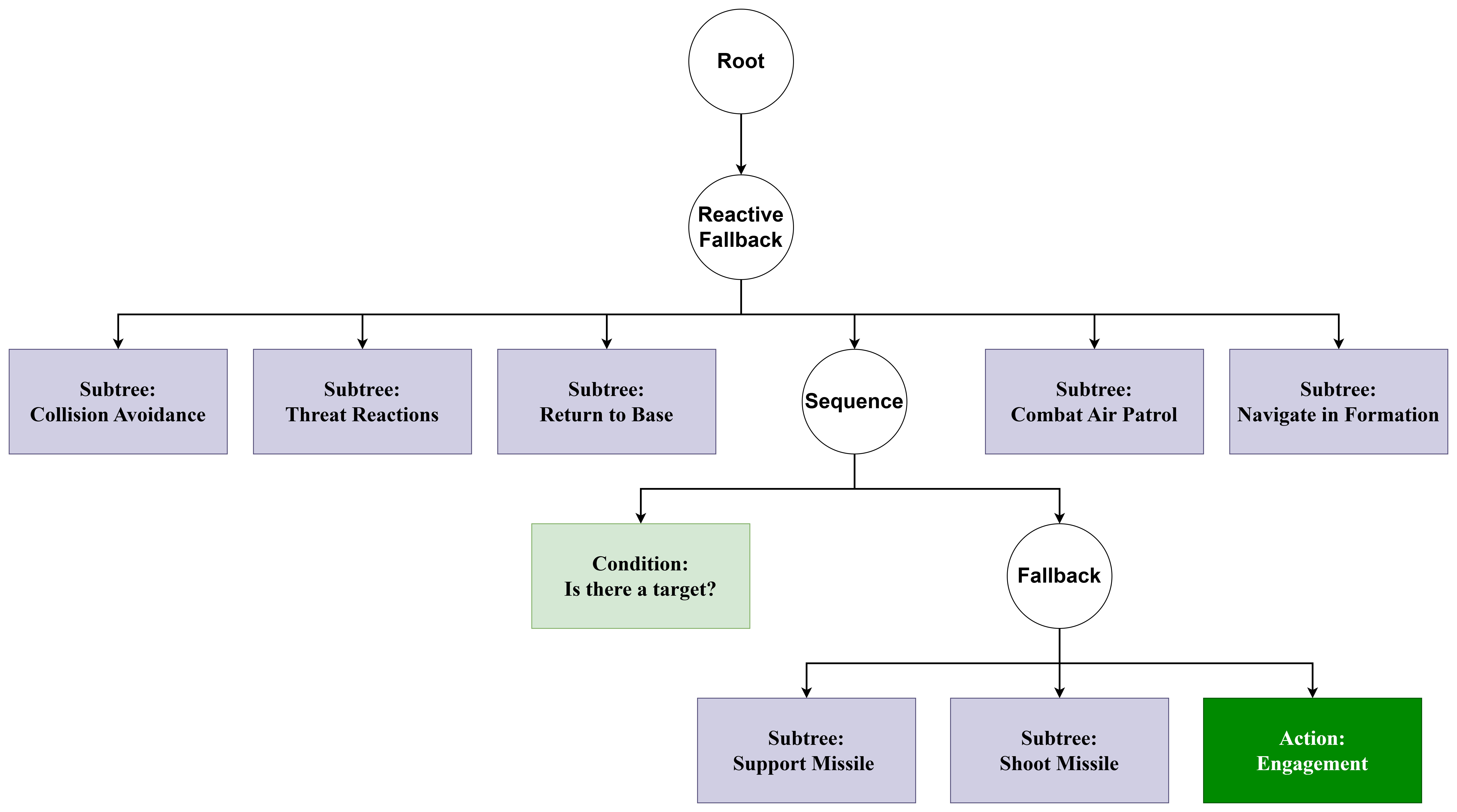}
\caption{Simplified Behavior Tree of the fighter agent. Only the engagement node, at the lower right, is replaced by the learned policy~\cite{dantas2025simulation}.}
\label{fig:btrl}
\end{figure}

We built AsaGym, a library that wraps ASA as a Gymnasium environment~\cite{towers2023gymnasium}, so standard RL implementations run without modification~\cite{raffin2019stable}. As shown in Figure~\ref{fig:btrl}, only the engagement node of the fighter BT is redirected to an external process, which exchanges observations and actions over a ZeroMQ socket using Protobuf messages. The other subtrees, covering collision avoidance, threat reactions, return to base, combat air patrol, and formation flight~\cite{costa2022formation}, keep running unchanged.

\begin{figure}[ht]
\centering
\includegraphics[width=0.64\textwidth]{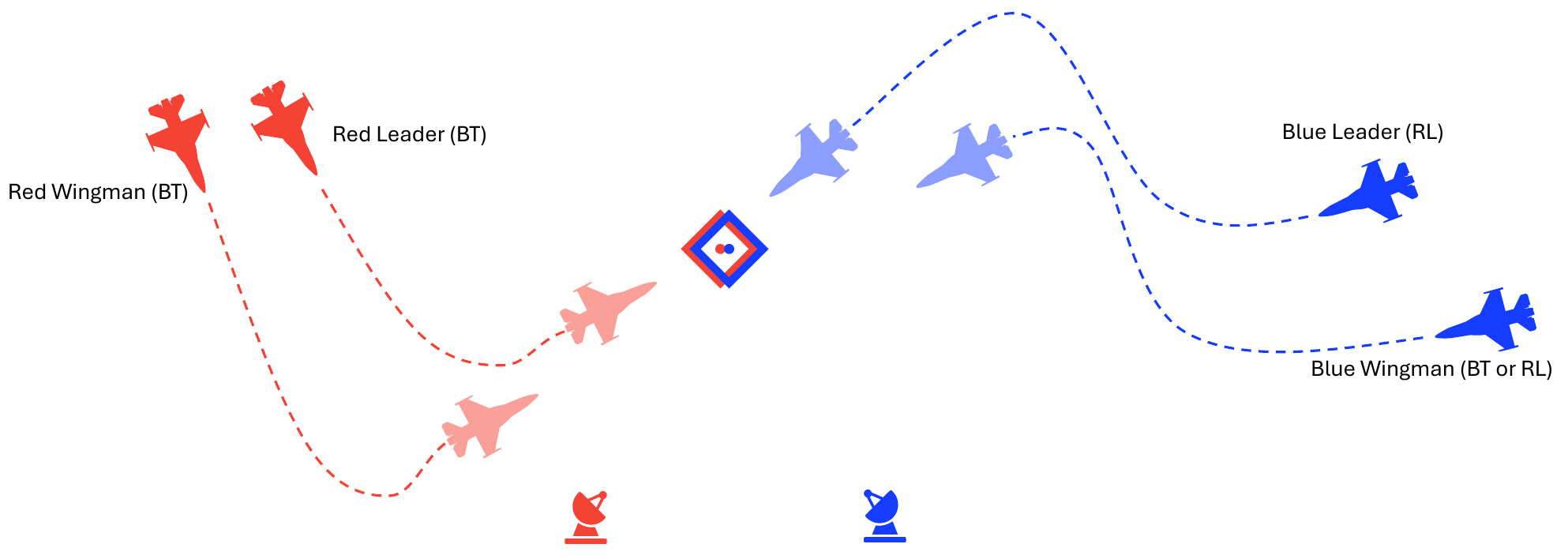}
\caption{Two-versus-two engagement. In Setup 1 only the blue leader is controlled by a learned policy; in Setup 2 both blue aircraft share a single policy. Each team defends a common point of interest and is supported by ground radar~\cite{dantas2025simulation}.}
\label{fig:scenario2v2}
\end{figure}

The task is a two-versus-two BVR engagement between F-16 models equipped with radar, radar warning receiver, datalink, and six active missiles~\cite{dantas2025asa}. Two setups are evaluated, shown in Figure~\ref{fig:scenario2v2}. In \textbf{Setup 1}, only the blue leader is controlled by a learned policy, while its wingman and both opponents follow expert-written engagement rules. In \textbf{Setup 2}, both blue aircraft share a single policy, each acting on its own state but receiving a shared observation of the whole scene. Episodes start head-on at 25,000 feet, with separation drawn uniformly from 20 to 60 nautical miles, and end when any aircraft is eliminated or a higher-priority BT node takes over. Altitude is held constant, so the engagement takes place in the horizontal plane.

\subsection{Observation, Action, and Reward}

The observation encodes the collective situational awareness of the engagement: for each of the four blue-red pairs and for the blue pair itself, it holds the magnitude difference of relative azimuths, the slant distance, and the airspeed difference, all normalized. This makes the policy invariant to where the engagement happens and to the absolute heading of the formation. The action is a continuous desired heading and airspeed, tracked by the underlying flight model, held for roughly five seconds by a frame-skipping wrapper so the agent commits to a maneuver instead of oscillating at the simulation rate.

The reward encodes what pilots call collective situational awareness. The \emph{relative azimuth} term rewards the blue team for pointing at the red team while the red team points away, summed over all pairs:
 
\begin{equation}
\text{reward}_{ra} = \sum_{i=1}^{n_B} \sum_{j=1}^{n_R} \frac{\left| ra(R_j,B_i) \right| - \left| ra(B_i,R_j) \right|}{n_B \, n_R \, ra_{\max}},
\end{equation}
where $n_B$ and $n_R$ are the number of blue and red aircraft, $B_i$ and $R_j$ are the $i$-th blue and $j$-th red agent, $ra(A,B)$ is the relative azimuth of $A$ with respect to $B$, and $ra_{\max} = 180^{\circ}$. A \emph{velocity} term repeats this angular advantage for the nearest opponent alone and scales it by the airspeed difference between the two aircraft, so an agent holding a favorable angle without the speed to exploit it earns less. It applies only when the angular advantage is positive, and is zero otherwise. The total reward is the sum of the two normalized terms.

The key design point is that the reward is summed over the team rather than computed per agent. In preliminary qualitative tests, not a controlled ablation, individualistic rewards such as minimizing the distance to an opponent or maximizing one's own angular advantage produced agents that pursued conflicting goals and left the formation uncovered, which motivated a team-level formulation that aligns each agent's objective with that of the formation.

\subsection{Results}

Four algorithms were compared using Stable-Baselines3 defaults~\cite{raffin2019stable}: Proximal Policy Optimization (PPO)~\cite{schulman2017proximal}, Advantage Actor-Critic (A2C)~\cite{mnih2016asynchronous}, Soft Actor-Critic (SAC)~\cite{haarnoja2018soft}, and Twin Delayed Deep Deterministic Policy Gradient (TD3)~\cite{fujimoto2018addressing}. The only changes were the rollout parameter (\texttt{n\_steps} for PPO and A2C, \texttt{train\_freq} for SAC and TD3) and, where applicable, the batch size, both set to 128. Each was trained five times with different seeds for 500,000 steps across 16 parallel environments.

\begin{figure}[ht]
\centering
\includegraphics[width=0.44\textwidth]{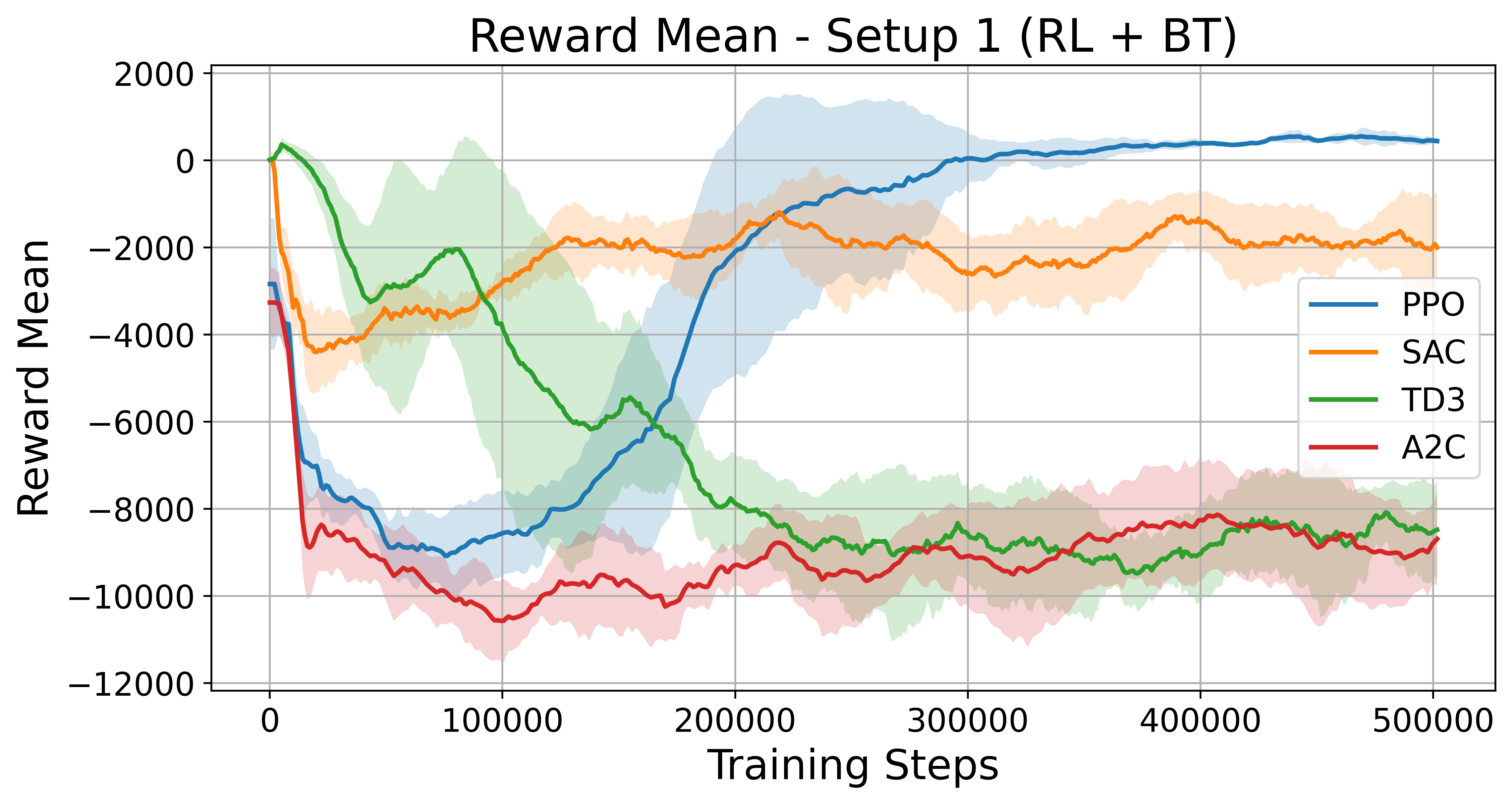}
\hspace{0.04\textwidth}
\includegraphics[width=0.44\textwidth]{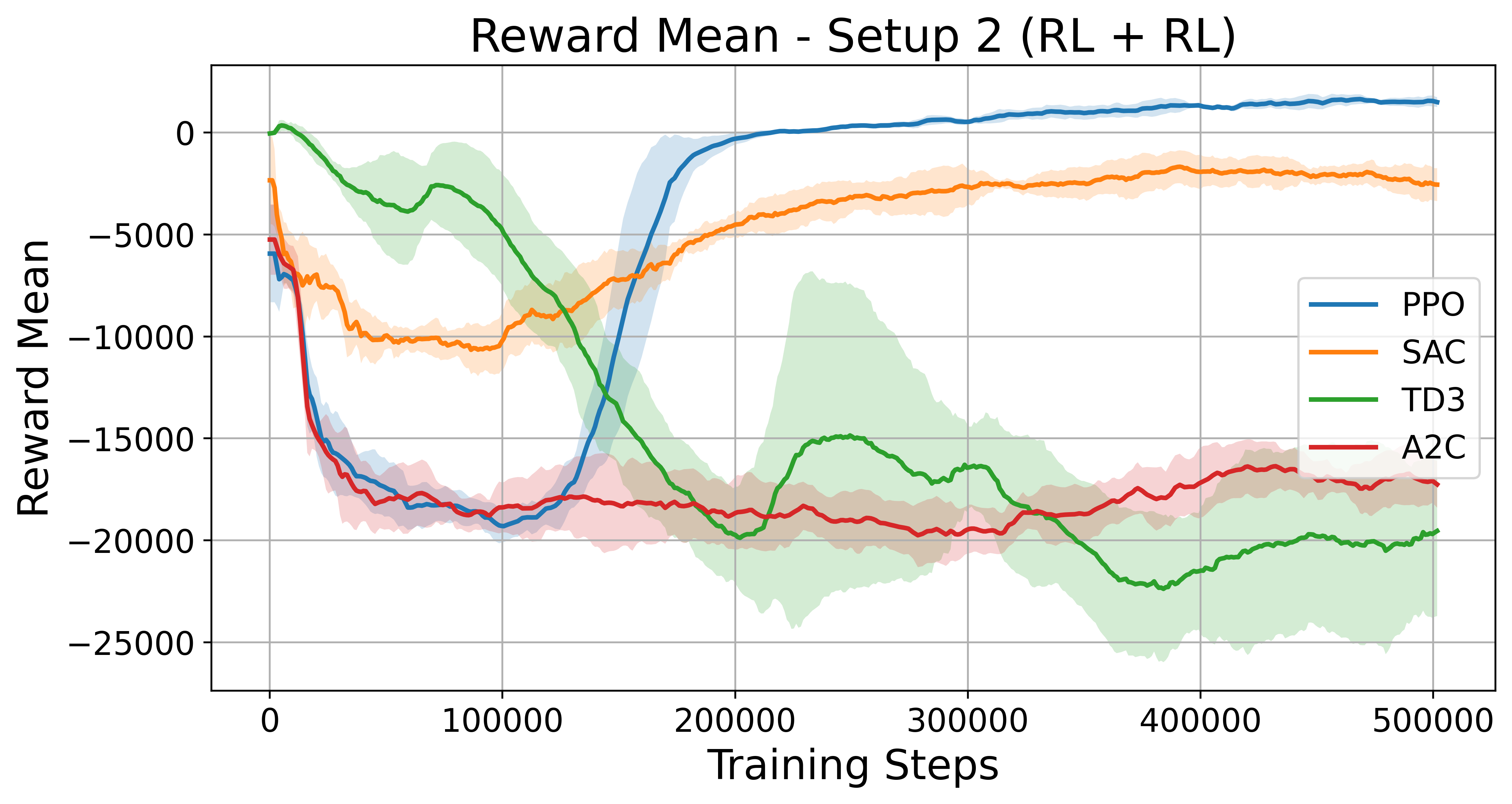}\\[2pt]
\includegraphics[width=0.44\textwidth]{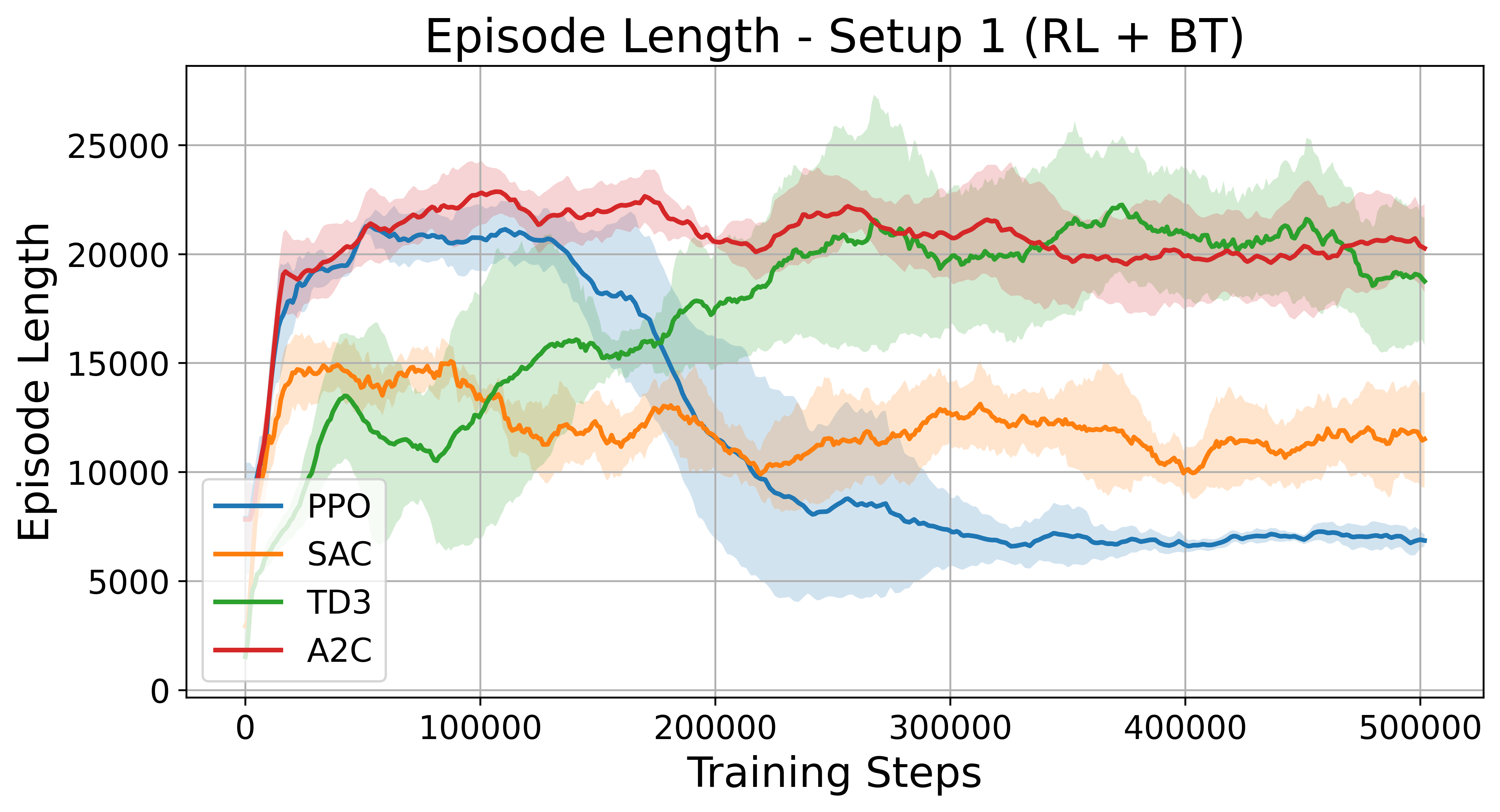}
\hspace{0.04\textwidth}
\includegraphics[width=0.44\textwidth]{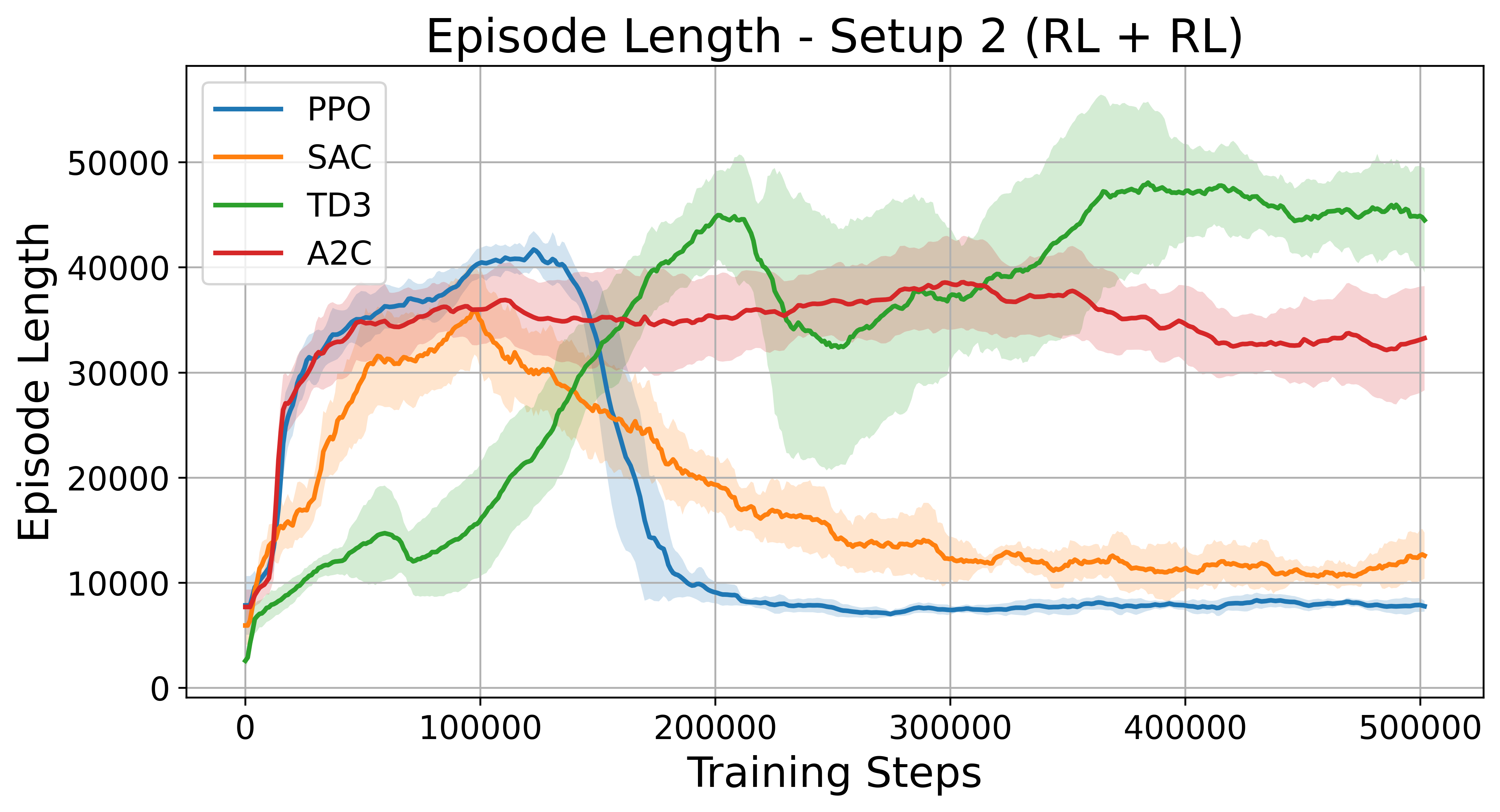}
\caption{Average episode reward (top) and mean episode length (bottom) over five seeds, in Setup 1 (left) and Setup 2 (right). The solid line is the mean and the shaded band the standard deviation~\cite{dantas2025simulation}.}
\label{fig:reward}
\end{figure}

The top row of Figure~\ref{fig:reward} shows the reward curves for both setups. PPO converges highest in both, and it stabilizes even faster in Setup 2, the harder task, which suggests that sharing one policy across two agents helped rather than hurt. SAC learns stably but settles lower. TD3 rises early and then degrades, consistent with favorable random choices rather than a learned policy, and A2C makes little progress.

Episode length, in the bottom row, separates the algorithms in the same order, with PPO and SAC stabilizing at the shortest episodes and A2C staying high throughout. Length alone is ambiguous, since an episode ends when any aircraft is eliminated, but paired with high reward, the short episodes of PPO and SAC mark decisive engagements rather than early losses.

\begin{table}[ht]
\centering
\scriptsize
\caption{Training time in hours, mean $\pm$ standard deviation over five seeds.}
\label{tab:drltime}
\begin{tabular}{lcccc}
\toprule
\textbf{Setup} & \textbf{PPO} & \textbf{SAC} & \textbf{TD3} & \textbf{A2C} \\
\midrule
Setup 1 (policy + BT)   & $6.77 \pm 0.57$ & $5.72 \pm 0.26$ & $4.74 \pm 0.24$ & $4.10 \pm 0.25$ \\
Setup 2 (shared policy) & $6.30 \pm 0.16$ & $5.02 \pm 0.31$ & $3.45 \pm 0.10$ & $3.10 \pm 0.14$ \\
\bottomrule
\end{tabular}
\end{table}

PPO was the most expensive to train (Table~\ref{tab:drltime}) but produced the best policies; Setup 2 trained faster because two learning agents double the experience per update.

\begin{figure}[tbp]
\centering
\includegraphics[width=0.68\textwidth]{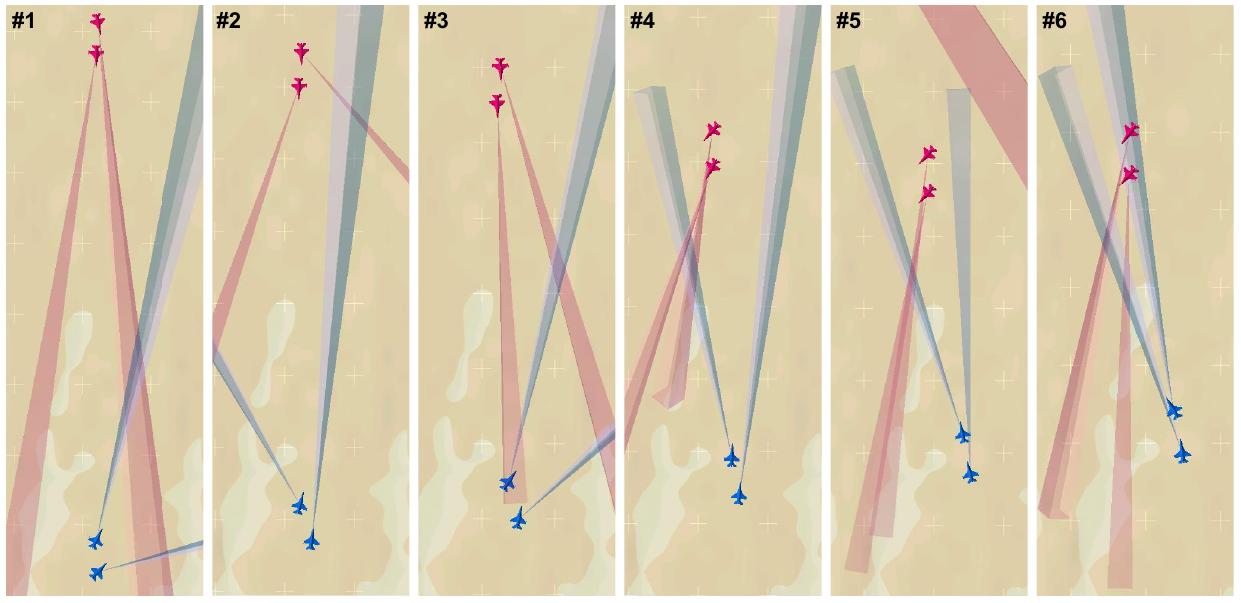}
\caption{Six frames from a typical engagement in Setup 2, showing how the two policy-controlled aircraft maneuver together to gain positional advantage~\cite{dantas2025simulation}.}
\label{fig:frames}
\end{figure}

Brazilian Air Force BVR fighter pilots, acting as subject matter experts, reviewed recorded PPO episodes through visual inspection. Figure~\ref{fig:frames} shows six frames from a representative Setup 2 episode. The experts identified behaviors they recognized from real training: mutual support, coordinated maneuvers, and adaptive targeting. The agents were not told to cooperate and the behavior emerged from a reward defined over the team.

\section{Learning Maneuvers from Human Pilots}
\label{sec:imitation}

Reward-driven learning works when the objective can be written down. Many maneuvers are instead defined by how an expert executes them, and for those, learning from demonstration fits better~\cite{dantas2024autonomous,dantas2025autonomous}.

\subsection{Task, Data, and Models}

We study the pop-up attack, an air-to-ground maneuver in which an aircraft climbs from low altitude, rolls, and dives onto a ground target while minimizing exposure. Holding the planned climb and dive angles while correcting for wind is what makes it hard to reproduce autonomously. The dataset contains 30 recordings flown by a Brazilian Air Force pilot in a six-degree-of-freedom F-16 simulator, all from the same starting point and trimmed to 39 frames.

Two state vectors are compared. The \emph{baseline} holds the seven quantities an instrument reports directly: altitude, pitch, roll, and yaw angles, radial angle, distance to target, and altitude difference. The \emph{full} configuration adds nine derived ones, the linear velocities, angular velocities, and accelerations, which must be estimated or computed rather than measured directly. The action vector is the pilot's own input: pitch, roll, and throttle. The baseline matters because it is what the model sees whenever those derived quantities are missing or their estimates are unreliable.

Behavior cloning used a multilayer perceptron (MLP), a long short-term memory (LSTM) network~\cite{hochreiter1997long}, and a gated recurrent unit (GRU) network~\cite{cho2014learning}, each with 128, 64, and 32 hidden units and a three-unit linear output. The recurrent models use a five-step sliding window with stride one. The 30 recordings were split into 70\% for training/validation and 30\% for testing, with five-fold cross-validation at the recording level, so windows from one recording stay in the same fold. Training used Adam with a learning rate of $10^{-4}$, batch size 64, and early stopping with a patience of 20 epochs. Final models were trained for the average number of epochs across folds, with results averaged over five seeds.

\subsection{Results}

Table~\ref{tab:il} reports the coefficient of determination ($R^2$) and the root mean squared error (RMSE), computed on the denormalized control inputs and averaged across the three commands; since these have different scales, RMSE serves to compare models rather than as a physical error. Two results stand out. First, the recurrent models retain far more explained variance when derived variables are removed: in $R^2$ the MLP falls from $0.726$ to $0.323$, while the GRU falls only from $0.971$ to $0.670$. RMSE is similar across all three models in the baseline, so the gain lies in tracking the shape of the maneuver rather than in absolute error. Temporal structure substitutes for explicit velocity and acceleration inputs, which matters whenever a robot acts on incomplete state. Second, the GRU matches the LSTM in accuracy with a simpler cell, since its two gates replace the three of the LSTM; in the full configuration it was also slightly faster, although within one standard deviation, which makes it a natural candidate for onboard execution.

\begin{table}[ht]
\centering
\scriptsize
\caption{Behavior cloning results, mean $\pm$ standard deviation over five seeds. Inference time is the time to predict the entire test set.}
\label{tab:il}
\begin{tabular}{llccc}
\toprule
\textbf{Model} & \textbf{Configuration} & \textbf{R\textsuperscript{2}} & \textbf{RMSE} & \textbf{Inference (ms)} \\
\midrule
MLP  & Baseline & $0.323 \pm 0.009$ & $8.850 \pm 0.015$ & $267.52 \pm 19.19$ \\
MLP  & Full     & $0.726 \pm 0.070$ & $1.977 \pm 0.054$ & $\mathbf{266.92 \pm 12.25}$ \\
LSTM & Baseline & $0.672 \pm 0.003$ & $8.708 \pm 0.059$ & $1432.22 \pm 122.63$ \\
LSTM & Full     & $0.970 \pm 0.001$ & $1.743 \pm 0.088$ & $1283.86 \pm 119.91$ \\
GRU  & Baseline & $0.670 \pm 0.003$ & $8.698 \pm 0.055$ & $1716.36 \pm 304.40$ \\
GRU  & Full     & $\mathbf{0.971 \pm 0.002}$ & $\mathbf{1.738 \pm 0.064}$ & $1206.61 \pm 65.94$ \\
\bottomrule
\end{tabular}
\end{table}

\begin{figure}[ht]
\centering
\includegraphics[width=0.96\textwidth, trim={0pt 10pt 0pt 20pt}, clip]{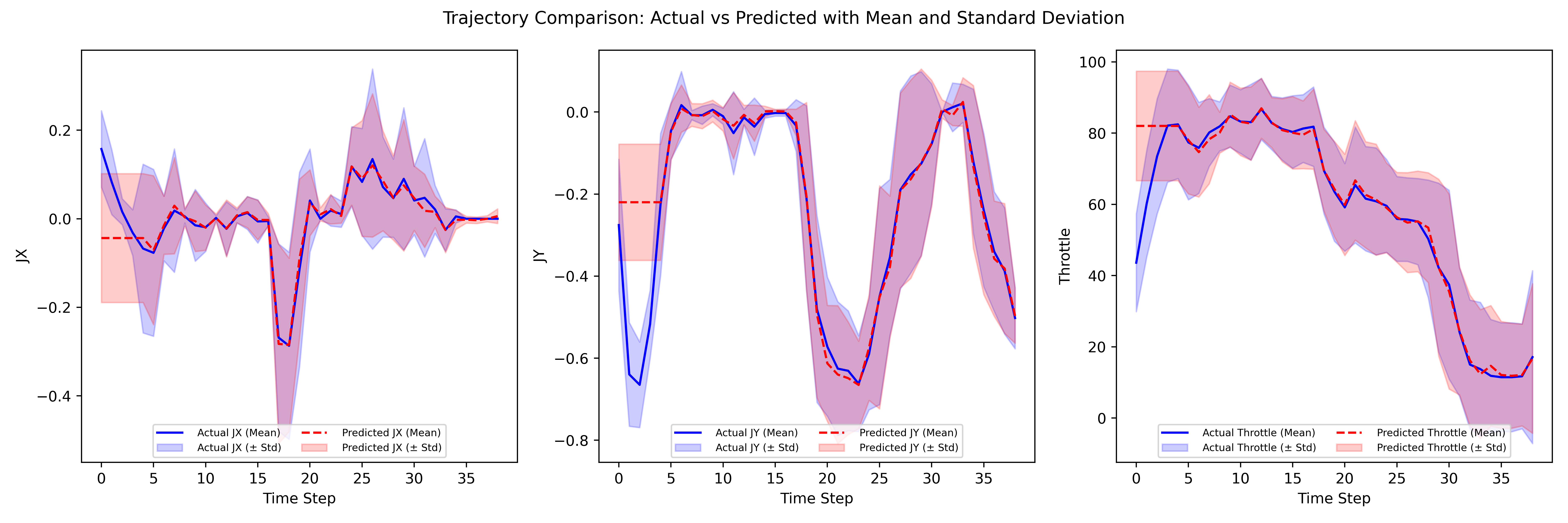}
\caption{Actual versus predicted pilot commands for pitch, roll, and throttle during the pop-up attack, using the GRU in the full configuration. Shaded areas show the standard deviation across recorded maneuvers and model predictions~\cite{dantas2025simulation}.}
\label{fig:traj}
\end{figure}

Figure~\ref{fig:traj} compares the predicted and actual commands. Roll is tracked most accurately, including during sharp transitions, while pitch slightly underestimates the amplitude in the most variable segments. Throttle follows the sustained phases well, with larger deviations occurring mainly during abrupt changes. Overall, prediction errors stay within the variability across the 30 pilot recordings, so the model captures the pilot's range of control behavior rather than any single flight.

\subsection{Generative Augmentation}

With only 30 demonstrations, a variational autoencoder~\cite{kingma2013auto} synthesized additional trajectories, smoothed with a moving average and reviewed by subject matter experts. Synthetic samples were added in amounts from 7 to 150.

The effect depended on how much real information the model already had, as Table~\ref{tab:synth} shows. In the baseline configuration the MLP gained most in $R^2$, rising to $0.455$ at 60 samples although its RMSE rose slightly, while the recurrent models changed little because they already extract the structure the extra samples would supply. In the full configuration synthetic data made every model worse, monotonically. Where real information is not missing, augmentation only introduces distribution mismatch.

\begin{table}[ht]
\centering
\scriptsize
\caption{Effect of synthetic data on R\textsuperscript{2}, mean $\pm$ standard deviation over five runs. Best mean per model and configuration in bold.}
\label{tab:synth}
\setlength{\tabcolsep}{3pt}
\begin{tabular}{llcccccc}
\toprule
& & \multicolumn{6}{c}{\textbf{Synthetic samples added}} \\
\cmidrule(l){3-8}
\textbf{Config.} & \textbf{Model} & \textbf{0} & \textbf{7} & \textbf{15} & \textbf{30} & \textbf{60} & \textbf{150} \\
\midrule
\multirow{3}{*}{Baseline}
 & MLP  & $0.323 \pm 0.009$ & $0.277 \pm 0.069$ & $0.312 \pm 0.066$ & $0.414 \pm 0.074$ & $\mathbf{0.455 \pm 0.053}$ & $0.439 \pm 0.043$ \\
 & LSTM & $0.672 \pm 0.003$ & $0.680 \pm 0.003$ & $0.679 \pm 0.005$ & $0.680 \pm 0.004$ & $0.680 \pm 0.005$ & $\mathbf{0.681 \pm 0.004}$ \\
 & GRU  & $0.670 \pm 0.003$ & $\mathbf{0.683 \pm 0.002}$ & $\mathbf{0.683 \pm 0.005}$ & $0.681 \pm 0.004$ & $0.680 \pm 0.007$ & $0.681 \pm 0.006$ \\
\midrule
\multirow{3}{*}{Full}
 & MLP  & $\mathbf{0.726 \pm 0.070}$ & $0.698 \pm 0.085$ & $0.616 \pm 0.094$ & $0.598 \pm 0.078$ & $0.472 \pm 0.140$ & $0.397 \pm 0.213$ \\
 & LSTM & $\mathbf{0.970 \pm 0.001}$ & $0.957 \pm 0.002$ & $0.951 \pm 0.002$ & $0.950 \pm 0.003$ & $0.944 \pm 0.004$ & $0.939 \pm 0.003$ \\
 & GRU  & $\mathbf{0.971 \pm 0.002}$ & $0.958 \pm 0.002$ & $0.954 \pm 0.004$ & $0.948 \pm 0.005$ & $0.943 \pm 0.005$ & $0.937 \pm 0.007$ \\
\bottomrule
\end{tabular}
\end{table}

\section{Discussion}
\label{sec:discussion}

Three lessons generalize beyond air combat. First, replacing one node of a behavior tree with a learned policy, rather than learning the whole controller, kept the system inspectable; this hybrid pattern applies to any robot that already has a working symbolic controller. Second, defining the reward over the team rather than the individual led to cooperative behavior without scripting it, relevant to multi-robot coordination wherever several vehicles share a task. Third, when demonstrations are few, temporal models tolerate missing state far better than feedforward ones, and synthetic data helps only when that state is genuinely incomplete. As for limitations, everything runs in high-fidelity simulation: AsaGym links the policy to ASA through a socket, which is not software- or hardware-in-the-loop testing, and behaviors were validated qualitatively by experts rather than against real flight data. The RL engagement is planar, so it does not yet support claims of three-dimensional autonomy, and the non-adaptive opponents risk overfitting the policy to a predictable adversary. Finally, comparisons are among algorithms and models within the framework; prior BVR agents run on different, mostly closed simulators, which prevented a direct quantitative comparison.

\section{Conclusion}
\label{sec:conclusion}

This paper summarized a doctoral thesis on simulation and machine learning for air combat. With a team-level reward, RL agents developed cooperative tactics that fighter pilots recognized from their own training; with 30 expert flights, temporal models reproduced a demanding maneuver even with incomplete state; and the study of generative augmentation clarified under which conditions synthetic data improves or degrades performance.

Future work on the RL side includes three-dimensional engagements, decentralized policies for larger formations, self-play against adaptive opponents, and a probability-of-kill term in the reward; on the imitation side, flights from more pilots and maneuvers, adversarial imitation, transformer-based models, and validation against real flight data. On both fronts, software- and hardware-in-the-loop testing is the next step toward deployment.

\section*{Acknowledgments}

This research was partially funded by the Funding Authority for Studies and Projects (FINEP), Finance Code 01.20.0196.00/2824/20.

\section*{Source Code}

Code is publicly available: AsaGym at \url{https://github.com/ASA-Simulation/asa-gym} and the imitation and generative learning models at \url{https://github.com/jpadantas/pop-up_attack_generative}. ASA and the recorded flight data are restricted due to their defense applications.

\bibliographystyle{sbc}
\bibliography{referencias}

\end{document}